# SANGRIA: Stacked Autoencoder Neural Networks with Gradient Boosting for Indoor Localization

Danish Gufran, Saideep Tiku, Sudeep Pasricha

*Abstract*— Indoor localization is a critical task in many embedded applications, such as asset tracking, emergency response, and real-time navigation. In this article, we propose a novel fingerprinting-based framework for indoor localization called SANGRIA that uses stacked autoencoder neural networks with gradient boosted trees. Our approach is designed to overcome the device heterogeneity challenge that can create uncertainty in wireless signal measurements across embedded devices used for localization. We compare SANGRIA to several state-of-the-art frameworks and demonstrate 42.96% lower average localization error across diverse indoor locales and heterogeneous devices.

## I. INTRODUCTION

Indoor localization involves the use of various technologies to determine the location of a device or person inside an enclosed space. Localization across indoor locales can serve several use cases, such as asset tracking, emergency response, and real-time indoor navigation [1]. As GPS signals are not able to penetrate buildings and other indoor environments, radio frequency signals such as Bluetooth low energy (BLE) and Wi-Fi are often utilized for localization [2]. One of the most promising methods for indoor localization involves measuring the received signal strength (RSS) of Wi-Fi signals, transmitted by Wi-Fi access points (APs), across multiple locations in an indoor floorplan. The measured RSS values at any indoor location forms a unique 'fingerprint' for that location that can then be used by a device to estimate location. This approach is popularly known as Wi-Fi RSS fingerprinting-based indoor localization [3].

In the fingerprinting approach, there are typically two phases: offline and online. During the offline phase, the RSS of Wi-Fi APs is captured at locations throughout the indoor space using one or more devices, such as smartphones and smartwatches. The resulting Wi-Fi RSS fingerprint data at each location is used to create a fingerprint database. A machine learning (ML) model, such as a neural network, can then be trained with the fingerprint database, such that the observed RSS fingerprint is the input, and the location is the output of the model. In the online phase, the trained ML model is deployed and used for localization. During this phase, a device measures the RSS of the Wi-Fi APs at an unknown indoor location, and the measured RSS values are sent to the ML model which in turn predicts the current location of the device.

There are several factors that can affect the accuracy of indoor localization frameworks [3]. Indoor environments are often highly cluttered, with many static and dynamic obstacles that cause unforeseeable wireless signal attenuation, as well as multipath fading and shadowing. These factors introduce noise in measured fingerprints, which can degrade the localization accuracy. Another major challenge arises due to heterogeneity in device characteristics, e.g., wireless radios that are used to transmit and receive Wi-Fi data, and software used to process the Wi-Fi signals, such as the algorithms used to filter out noise and errors [4]. Because of device heterogeneity, two different devices may end up capturing very different RSS values at the same indoor location. This phenomenon leads to reduction in indoor localization accuracy across devices. Consequently, it is crucial to devise indoor localization frameworks that can not only cope with uncertainties across indoor environments, but also factor in the impact of user device heterogeneity.

In this article, we present SANGRIA, a novel neural network-based Wi-Fi fingerprinting framework for indoor localization that is specifically designed to address the device heterogeneity challenge. Our approach combines the benefits of a stacked autoencoder neural network and gradient boosted trees, resulting in a highly accurate and robust approach for indoor localization. Our novel contributions are as follows:

- We propose a novel and uniquely tailored data augmentation technique for handling device heterogeneity;
- We propose a framework with gradient boosting algorithms to minimize indoor localization error;
- Through capturing fingerprints from several smartphones across diverse buildings, we establish real-world benchmarks to compare SANGRIA's localization accuracy against several state-of-the-art localization frameworks.

## II. RELATED WORK

Wi-Fi RSS fingerprinting is a particularly promising method for indoor localization due to the ubiquitous use of Wi-Fi capable devices by users and availability of Wi-Fi APs in indoor environments [2]. Several studies have applied ML algorithms to learn patterns in Wi-Fi RSS data and provide accurate localization predictions [3].

The work in [5] employs k-nearest-neighbors (KNN) and proposes an RSS averaging technique to improve localization accuracy but fails to show consistent results across buildings. In [6], random forests (RF) and support vector machines (SVM) are enhanced by incorporating principle component analysis (PCA) for feature extraction. The work in [7] combines gaussian process classifiers (GPC) with an augmentation strategy to deal with noisy RSS data, but for a simpler localization problem with lower resolution. Several recent ML techniques have used gradient-boosted trees (GBT) for indoor localization [8]-[10]. The GBT algorithm in [8] is applied on BLE RSS data. However, Wi-Fi RSS is more effective for localization compared to BLE RSS due to its wide availability, high accuracy, low cost, and longer range compared to BLE [9]. The works in [10] and [11] employ a simple GBT model [10] and XgBoost (extreme gradient boosting) [11] to improve localization accuracy. However, these works do not address challenges due to the variations caused by device heterogeneity.

Neural networks, e.g., deep neural networks (DNN) [12] and convolutional neural networks (CNN) [13] have also been explored to improve Wi-Fi RSS based indoor localization. The framework in [12] aims to address device heterogeneity but shows higher localization errors compared to the approaches in [5]-[11]. The work in [13] demonstrates lower localization errors than [12] but does not consider device heterogeneity. The work in [14] adapts multi-head attention neural networks for localization but shows higher worst-case localization errors for certain testing devices, lacking strategic augmentation.

To mitigate device heterogeneity, augmenting training data plays a critical role in inhibiting overfitting. A few prior works have used augmentation techniques for indoor localization. In [12], generative adversarial networks (GANs) are used as an augmentation module, while [7] employs stacked autoencoders (SAEs), and [8] employs a simple neural network.

After thoroughly analyzing the prior works, we devise SANGRIA that combines a carefully tailored augmentation technique based on Stacked Autoencoder (SAE) [7] with a powerful gradient boosting algorithm [10], [11] to achieve high accuracy in a resource-efficient manner.

## III. SANGRIA Framework

### A. Overview

Our proposed SANGRIA framework (figure 1) consists of two main components: a greedy layer-wise stacked autoencoder (custom SAE) and a categorical gradient boosting algorithm.

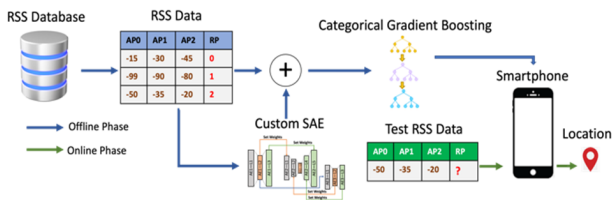

Figure 1: Implementation overview of the SANGRIA framework

In the offline phase, SANGRIA involves data collection, data pre-processing, data augmentation, and ML model training and deployment. The data collection and pre-processing phases are used to generate an RSS fingerprint database. A custom stacked autoencoder (SAE) model is tailored to augment fingerprints from the database such that an ML model can produce accurate predictions despite device heterogeneity. The collected data prior to augmentation and the augmented data are concatenated to diversify the RSS input data. The concatenated data is then used to train the categorical gradient boosting ML model, to learn the relationship between RSS measurements and indoor locations. Finally, this trained model is deployed on the smartphone. The offline phase is typically performed once, and the resulting model is used for multiple localization estimates in the online phase. The accuracy of the location estimates is determined by computing error using the Euclidian distance:

$$Distance = \sqrt[2]{(X_2 - X_1)^2 + (Y_2 - Y_1)^2 + (Z_2 - Z_1)^2} \quad (1)$$

where $(X_1, Y_1, Z_1)$ and $(X_2, Y_2, Z_2)$ are the values of the ground truth and predicted location coordinates, respectively.

### B. Data Collection, Preprocessing, and Database Creation

For any indoor environment of interest, we collect Wi-Fi RSS values in dBm (decibel-milliwatts). Each RSS value ranges between –100dBm to 0dBm, where –100 indicates no signal and 0 indicates a full (strongest) signal. We collect RSS values from multiple (six in our experimental analysis) smartphones from different manufacturers, with different Wi-Fi chipsets. The collected data is pre-processed by normalizing and splitting it into train and test datasets for subsequent use with the ML algorithm. We normalize all RSS values between the range of 0 (–100dBm) to 1 (0dBm). The resulting normalized vector of RSS values for each location reference point (RP) represents a fingerprint and is stored in the RSS database. To evaluate the ML model, we split the collected RSS data into training and testing datasets.

### C. Greedy Layer-Wise Stacked Autoencoder

To augment the collected data before training, we design a custom stacked autoencoder (SAE) neural network model. Our SAE model is composed of multiple autoencoders (AE), as shown in figure 2, which are trained one layer at a time using a greedy layer-wise training approach.

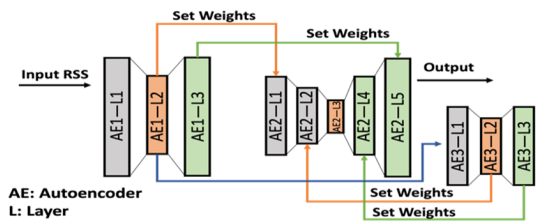

Figure 2: Greedy layer-wise stacked autoencoder (SAE) architecture with Encoder (gray), Decoder (green), and bottleneck (orange) layers.

An AE neural network consists of three main parts: an Encoder, a Decoder, and a Bottleneck layer. Each of these parts is composed of one or more fully connected layers with varying number of neurons. The encoder layers are used to compress the input data (dimensionality reduction) and the decoder layers are used to reconstruct a noisy representation of the input data, back to its original dimension. The bottleneck layer is a representation of the input data at its lowest dimension, representing important features in the data.

The proposed SAE consists of three AEs stacked together with each having its own fully connected layers, as shown in figure 2. AE1 and AE3 are trained layer-wise, where the output of the second layer (L2) of AE1 is fed as input to AE3. AE2 receives the trained weights from AE1 and AE3 and is positioned as shown in figure 2. The SAE is trained in a greedy layer-wise manner to minimize reconstruction error between the input and output data. The advantage of this approach is that it allows the network to learn a good representation of the input data at each layer before moving on to the next one, rather than trying to optimize all the layers together. This can make the training process more stable and can lead to better performance.

SAE is trained on the RSS database to generate synthetic RSS measurements, simulating variations across devices that are not overly biased towards the device used for training, resulting in better generalization and more accurate predictions.

### D. Categorical Gradient Boosting

Gradient-boosted trees (GBTs) are a powerful type of ML model that involve training a series of decision trees and combining their predictions to create a stronger overall model.

Instead of predicting the target variable, each decision tree in a gradient boosting model is trained to predict the residual error of the previous tree, which helps minimize the overall error during training. There are several types of GBT models, including CatBoost, XgBoost, and LightGBM. We chose to use CatBoost for our problem because it is better suited for dealing with categorical variables and does not require additional pre-processing steps, making it a more efficient choice. Additionally, its symmetric trees can provide more accurate predictions in cases with a large number of categorical features, such as the number of unique visible Wi-Fi APs in our indoor localization problem. To train our model, we iteratively update each decision tree to the residual error of the previous iteration, as per the mathematical representation in the equations below:

$$F(x) = T(\Sigma W_i * H_i(x)) \quad (2)$$
$$F(x) = F(x-1) + H(x) \quad (3)$$

where $H_i(x)$ is the prediction made by the $i^{th}$ decision tree, $W_i$ are the weights assigned to the $i^{th}$ tree, and $T$ is monotonic transformation applied to the combined predictions. The monotonic transformation is key as it retains the order of the RSS fingerprint for each tree (changes in the fingerprint order can lead to higher localization errors). The weights $W_i$ are representative of the learned features from the RSS fingerprint. $H(x)$ is the prediction made by the decision tree at iteration $x$. $F(x-1)$ is the prediction made by the model after $x-1$ iterations. $F(x)$ is the final prediction (location of the RSS fingerprint) made by the model after $x$ iterations.

## IV. EXPERIMENTS

### A. Experimental Setup

We collected fingerprint data from five buildings with different salient features on the Colorado State University campus. The buildings have path lengths varying between 60 to 80 meters and contain 80 to 180 Wi-Fi APs. RP locations for each fingerprint are one meter apart. Building 1 has wooden furniture and concrete walls, Building 2 has heavy metallic laboratory equipment, Building 3 and 4 have a mix of wood, concrete, and smaller metallic equipment and computers, and Building 5 has wide open spaces.

To account for device heterogeneity, we selected six smartphones: BLU Vivo 8 (BLU), HTC U11 (HTC), Samsung Galaxy S7 (S7), LG V20 (LG), Motorola Z2 (MOTO), and OnePlus 3 (OP3). During the training phase, we collected data from a single smartphone per building floorplan. We ensured that each smartphone captured between 61 to 78 unique RPs per floorplan, with 78 to 339 visible APs. We also took five samples per RP for training and one sample per RP for testing. In total, we collected 1606955 data points for our training dataset, which includes all floorplans. To further improve the size and quality of our training dataset, we utilized the custom SAE for data augmentation. The SAE effectively doubles the size of our training dataset, resulting in a total of 3213910 data points for training. We collected the RSS fingerprints in a real-world scenario and considered building floorplans with multiple salient features, including external obstacles that can add noise to the collected RSS data. Moreover, we collected training and testing data during normal working hours of the day to incorporate environmental factors such as human interference, dynamic obstacles, attenuation, fading, shadowing, and so on.

A detailed description of the number of trainable parameters in the proposed SAE within SANGRIA is shown in table 1. We enhanced the baseline CatBoost algorithm from the library in [15]. To fine-tune CatBoost within SANGRIA, we set number of iterations to 50, with a total depth in trees to 7, learning rate of 0.1, and L2 leaf regularization to 5 to prevent overfitting. In the next section we present experimental results of SANGRIA's performance comparison against the state-of-the-art.

TABLE 1: Trainable parameters for each stack in the SAE in SANGRIA

| AE1-L1 | AE1-L2 | AE1-L3 | AE2-L1 | AE2-L2 | AE2-L3 |
|--------|--------|--------|--------|--------|--------|
| 14878  | 10414  | 14964  | 10414  | 5103   | 3570   |
| AE2-L4 | AE2-L5 | AE3-L1 | AE3-L2 | AE3-L3 | Total  |
| 7388   | 14964  | 7289   | 5103   | 7388   | 101475 |

### B. Experimental Results

To evaluate the performance of SANGRIA, we select seven state-of-the-art indoor localization frameworks for comparison: KNN [5], RF [6], SVM [6], GPC [7], and DNN [12] that target localization with device heterogeneity resilience; and XgBoost [11] and CNN [13] which aim to aggressively minimize indoor localization error but do not address device heterogeneity.

We collect localization errors for each train-test combination across all paths. The results are plotted in figure 3 by averaging the localization errors across all floorplans. From figure 3, we can clearly observe low errors for SANGRIA compared to other frameworks. SANGRIA shows 28% to 66% improvements in mean, and 42% to 74% maximum error reduction, compared to the other frameworks. This is an indication of SANGRIA's excellent generalizability across all devices.

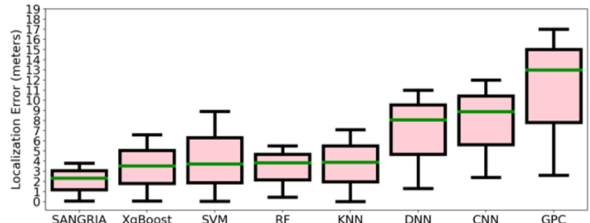

Figure 3: Min (lower whisker), Mean (green bar), and Max (upper whisker) error across all comparison frameworks for indoor localization.

To further analyze the individual device-wise performance of each framework, we construct a heatmap for the top four best performing frameworks, as shown in figure 4. The error (in meters) for each cell is the average for that configuration (train-test combination) across all five building floorplans. From figure 4, we see that SANGRIA improves the worst-case localization errors compared to the other frameworks. For example, in the LG-MOTO and LG-OP3 configurations, we observe higher errors in SVM and XgBoost, but these worst-case errors are greatly reduced for SANGRIA. Similarly, in the BLU-OP3 configuration, we observe high errors in RF and these errors are reduced for the rest of the frameworks with SANGRIA reporting the lowest error.

We have open sourced our dataset [17] to aid reproducibility. To further evaluate the effectiveness of SANGRIA, we trained the proposed ML models using another publicly available dataset: the UJI dataset [16] for indoor localization. The dataset contains Wi-Fi RSS measurements from 520 APs distributed across a three-story building. Each of the 520 Wi-Fi APs was equipped with one receiver and four antennas, and data was collected using 20 Wi-Fi cards. Figure 5 shows the mean localization errors of each model trained and tested on the UJI dataset. Notably, SANGRIA has the lowest localization errors, surpassing its closest competitor XgBoost by 11.4%.

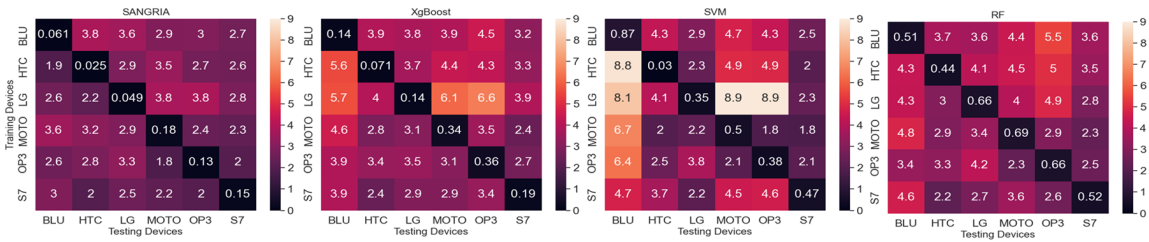

Figure 4: Experimental results for the top four best performing frameworks with average errors (meters) across all floorplans.

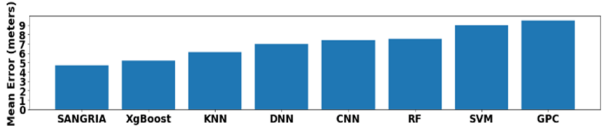

Figure 5: Mean localization errors across all frameworks using UJI dataset

SANGRIA's superior localization capabilities are to a large extent due to the uniquely tailored greedy-layer wise SAE. The other contributing factor is the excellent feature extraction ability of the CatBoost algorithm. CatBoost has several advantages over other algorithms due to its ability to handle categorical features, such as number of visible APs and MAC (Media Access Control address) ID of each Wi-Fi AP. In contrast, other algorithms such as DNN, CNN, SVM, RF, KNN, and GPC require the categorical features to be transformed into numerical form. CatBoost also can better reduce overfitting, which is a common problem in GBT algorithms. It does this by randomly selecting a subset of features to use at each iteration of the boosting process, making it a better choice than XgBoost.

### C. Effect of the Stacked Autoencoder

In figure 6, we observed a significant drop in mean localization errors for all frameworks when using SAE: 6.82% lower mean error for CatBoost, 17.64% lower mean error for XgBoost, 20.12% lower mean error for SVM, 8.16% lower mean error for RF, 2.64% lower mean error for KNN, 37.94% lower mean error for DNN, 47.34% lower mean error for CNN, and 87.59% lower mean error for GPC.

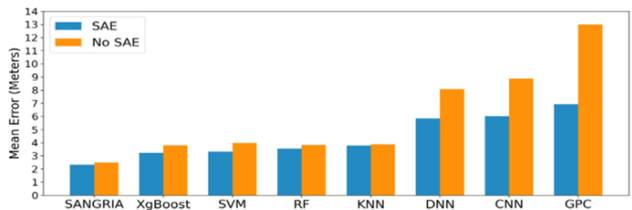

Figure 6: Bar plot comparing average errors of all frameworks with SAE (Blue) and without SAE (Orange)

TABLE 2: Avg. inference time (milliseconds) across devices and frameworks

|      | SANGRIA | XgBoost | SVM   | RF  | KNN  | DNN   | CNN | GPC    |
|------|---------|---------|-------|-----|------|-------|-----|--------|
| BLU  | **20**  | 30      | 400   | 90  | 30   | 630   | 680 | 1960   |
| HTC  | **10**  | 20      | 110   | 80  | 20   | 540   | 630 | 1830   |
| LG   | **20**  | 30      | 200   | 80  | 10   | 550   | 630 | 1860   |
| MOTO | **10**  | 10      | 190   | 70  | 20   | 540   | 610 | 1820   |
| OP3  | **10**  | 20      | 280   | 80  | 30   | 540   | 620 | 1810   |
| S7   | **10**  | 10      | 180   | 80  | 20   | 510   | 610 | 1740   |
| Avg. | **13.3**| 20      | 226.6 | 80  | 21.6 | 551.6 | 630 | 1836.6 |

### D. Model Inference Time Latency

Low latency in the online phase is important to achieve practical indoor localization framework deployment. Table 2 shows the measured latencies in milliseconds for each framework on smartphones. SANGRIA has the fastest inference time with an average of ~13.3ms, followed by KNN with ~21.6ms, and XgBoost with ~20ms. GPC has the highest latency due to its probabilistic computation.

## V. CONCLUSION

In this paper, we present SANGRIA, a novel neural network-based solution that is resilient towards device heterogeneity during indoor localization. Experimental results show a 42.96% reduction in average localization error and an average latency of 13.3 milliseconds, making it a reliable and practical approach for device-invariant indoor localization.